\pdfoutput=1

\documentclass[11pt]{article}

\usepackage{naacl2021}

\usepackage{times}
\usepackage{latexsym}

\usepackage[T1]{fontenc}

\usepackage[utf8]{inputenc}

\usepackage{microtype}

\usepackage{placeins}
\usepackage{amsfonts}
\usepackage{amsmath}

\usepackage{float}
\usepackage[utf8]{inputenc} 
\usepackage[T1]{fontenc}    
\usepackage{hyperref}       
\usepackage{url}            
\usepackage{booktabs}       
\usepackage{amsfonts}       
\usepackage{nicefrac}       
\usepackage{microtype}      
\usepackage{tcolorbox}
\usepackage{adjustbox}
\usepackage{multirow}
\usepackage[draft]{minted}
\usepackage{array}
\usepackage{siunitx}
\setminted{fontsize=\tiny}
\definecolor{bg}{HTML}{282828}
\usepackage{pmboxdraw}
\usepackage{algorithmic}
\usepackage{listings}
\usepackage[linesnumbered,ruled,vlined]{algorithm2e}
\def\BibTeX{{\rm B\kern-.05em{\sc i\kern-.025em b}\kern-.08em
    T\kern-.1667em\lower.7ex\hbox{E}\kern-.125emX}}

\begin{document}

\lstdefinestyle{myJavaStyle}{
  frame=tb,
  float=*,
  language=java,
  aboveskip=3mm,
  belowskip=3mm,
  showstringspaces=false,
  columns=flexible,
  basicstyle={\small\ttfamily},
  numbers=none,
  numberstyle=\tiny\color{gray},
  keywordstyle=\color{blue},
  commentstyle=\color{dkgreen},
  stringstyle=\color{mauve},
  frame=single,
  breaklines=true,
  breakatwhitespace=true,
  tabsize=3,
}

\title{Generating Bug-Fixes Using Pretrained Transformers}

\author{
  Dawn Drain \\
  Microsoft\\
  dawn.drain@microsoft.com\And
  
  Chen Wu \\
  Microsoft\\
  wu.chen@microsoft.com\AND 
  
  Alexey Svyatkovskiy \\
  Microsoft\\
  alsvyatk@microsoft.com\And
  
  Neel Sundaresan \\
  Microsoft\\
  neels@microsoft.com
  }

\date{April 2021}

\maketitle

\begin{abstract}
Detecting and fixing bugs are two of the most important yet frustrating parts of the software development cycle. Existing bug detection tools are based mainly on static analyzers, which rely on mathematical logic and symbolic reasoning about the program execution to detect common types of bugs. Fixing bugs is typically left out to the developer. 
In this work we introduce \texttt{DeepDebug}: a data-driven program repair approach which learns to detect and fix bugs in Java methods mined from real-world GitHub repositories. We frame bug-patching as a sequence-to-sequence learning task consisting of two steps: (i) denoising pretraining, and (ii) supervised finetuning on the target translation task.
We show that pretraining on source code programs improves the number of patches found by 33\% as compared to supervised training from scratch, while domain-adaptive pretraining from natural language to code further improves the accuracy by another 32\%. We refine the standard accuracy evaluation metric into non-deletion and deletion-only fixes, and show that our best model generates 75\% more non-deletion fixes than the previous state of the art. In contrast to prior work, we attain our best results when generating raw code, as opposed to working with abstracted code that tends to only benefit smaller capacity models. 
Finally, we observe a subtle improvement from adding syntax embeddings along with the standard positional embeddings, as well as with adding an auxiliary task to predict each token's syntactic class. Despite focusing on Java, our approach is language agnostic, requiring only a general-purpose parser such as tree-sitter\footnote{https://tree-sitter.github.io/tree-sitter/}.

\end{abstract}


\section{Introduction}
Early results in automated bug-patching assumed a set of test functions and took the approach of generating patches until one passed the test suite. The edits were operationalized as abstract-syntax tree (AST) manipulations such as deleting, inserting, or swapping nodes. The seminal work GenProg \cite{genprog} combined these operations with a genetic algorithm and claimed to fix 55 out of 105 bugs in 5.1m LOC across eight projects with over 10k test cases. However, a followup paper \cite{Qi} found that 53 out of the 55 claimed fixes were merely test-suite adequate, and that most of them were generated simply using the deletion operator.

More recent work has explored patch generation using a more stringent criterion: produce exactly the same fix as the developer who originally  fixed the bug. We use the Java dataset from \textit{Patches in the Wild} \cite{Tufano}, which is comprised of short methods extracted from commits whose messages indicate they are bug-fixing. \textit{Copy That!} \cite{panthaplackel2020copy} applies a novel span-copying mechanism on this dataset, achieving impressive top-1 accuracy. SequenceR \cite{sequencer} both narrows and expands on the \textit{Patches in the Wild} dataset by focusing on one-line changes only, while also providing additional context beyond the buggy method by including other methods' signatures. This extended context gives a 15\% relative boost. All three approaches see large gains from copying, which is sensible given that the fixed method has a large overlap with the buggy method. ENCORE \cite{encore} also looks at one-line changes, and in more languages, although they give only the buggy line as input and do not provide any further context. 


For certain bug-types it is possible to generate millions of synthetic bugs to train on. Devlin et al. \cite{Devlin} train an RNN on Python to fix incorrect comparison operators, the mistaken use of ``is'' vs. ``is not'', variable misuse, and forgotten ``self'' accessors. Overall, they achieve 86\% accuracy on synthetic bugs and 41\% on real-life bugs. Kanade et al. \cite{cuBERT} pretrain a BERT-style model ``CuBERT'' on a larger Python dataset and then finetune on a related suite of synthetic bugs, achieving over 90\% accuracy. 

In the broader field of NLP, our task is closest in spirit to grammatical error correction (GEC). Drawing from the CoNLL GEC leaderboard \cite{ng2014conll}, we see several different approaches using transformers. Kiyano et al. see large gains from data augmentation via noisy backtranslation, in addition to more standard techniques like injecting spelling errors, applying left-to-right reranking, and using a sentence-level error detection task \cite{Kiyono_pseudo_data}. Kaneko et al. use a seq2seq model where the input embeddings are concatenated with the embeddings produced by a BERT finetuned on grammatical error detection (detecting whether a given token is faulty) \cite{Kaneko_ged}. Zhao et al. \cite{Zhao} apply a copy-mechanism, in addition to injecting artificial noise and using an auxiliary token-labeling task. Awasthi et al. \cite{awasthi2019parallel} pursue an encoder-only approach, which maps token $z_i$ to the edit operation to perform at the $i^{th}$ location. They see large speed gains, and also investigate applying their model iteratively.

Expanding more broadly, still, various task-agnostic pretraining approaches like BERT, BART, and T5 \cite{Devlin, BART, T5} have seen large performance gains on composite benchmarks like GLUE \cite{wang2018glue}. These models are typically pretrained using a denoising objective on a large corpus of synthetically corrupted text. 

We retain the scope of work trying to fix generic bugs while taking advantage of techniques for pretraining powerful sequence-to-sequence transformers. We first treat raw code as text and use a span-masking objective to pretrain a 400 million parameter encoder-decoder transformer on 67k Java repositories mined from GitHub before finetuning on the \textit{Patches in the Wild} benchmark.  More specifically, we consider three pretraining experiments: pretrain on Java only; finetune directly from the strong baseline BART which was pretrained on English; and further pretrain on Java with a warmstart from BART. Pretraining on Java alone improves the number of patches found by a third, from 1049/12380 to 1392/12380, while pretraining with a warmstart further improves by another third to 1839/12380. We refine the standard top-1 accuracy metric used for this benchmark into non-deletion and deletion-only fixes, and show that our best \texttt{DeepDebug} model generates 75\% more non-deletion fixes than the previous state of the art. In contrast to prior work, we attain our best results when generating raw code, as opposed to working with abstracted code that tends to benefit smaller models. Finally, we see a subtle improvement from using the formal grammar unique to programming languages. Specifically, we experiment with adding syntax embeddings along with the standard positional embeddings, as well as with adding an auxiliary task to predict each token's syntactic class.

\section{Problem overview}
The task of patch generation is as follows: given a buggy Java method (found by mining commit messages), produce the exact same sequence of tokens as the developer who committed the fix. 
\begin{figure}[h]
\vspace{-0.2cm}
    \centering
\begin{adjustbox}{width=.48\textwidth}
\footnotesize
\begin{tabular}{c}
\toprule
  Example Found Fix \\
\midrule
\begin{minipage}[t]{0.45\textwidth}
\begin{minted}[escapeinside=||]{java}
double a_ods_light_Detected() {
    double l_return = 0.0;
    if ((v_sensor_ods) != null) {
        v_sensor_ods.getLightDetected();
    }
    return l_return;
}
\end{minted}
\end{minipage}\\
\midrule
\begin{minipage}[t]{0.45\textwidth}
\begin{minted}[escapeinside=||]{java}
double a_ods_light_Detected() {
    double l_return = 0.0;
    if ((v_sensor_ods) != null) {
        l_return |\colorbox{green}{=}| v_sensor_ods.getLightDetected();
    }
    return l_return;
}
\end{minted}
\caption{Our model makes sure to capture the value returned by the \texttt{getter} method.}
\end{minipage}\\
\end{tabular}
\end{adjustbox}
\vspace{-0.4cm}
\label{fig:first_fix}
\end{figure}

During finetuning, we train our sequence-to-sequence transformer so as to minimize the expected log-probability of generating the fix given the buggy method. Formally, let $B = \{b_1,\dots,b_n\}$ and $C = \{c_1,\dots,c_n\}$ be the sets of buggy and correct functions respectively, where we can tokenize $c_i = c_{i,1},\dots,c_{i,k_i}$. Let $\theta$ denote the model parameters. The objective is thus:
\vspace{-0.4cm}

\begin{equation*}
\begin{split}\text{argmin}_\theta\mathbb{E}_{b_i \in B} \text{log} & [P(c_{i,1} | b_i; \theta)\cdot P(c_{i,2} | b_i,c_{i,1}; \theta)
 \\ & \cdots P(c_{i,k_i} | b_i,c_1,\dots,c_{k_i-1}; \theta)].
\end{split}
\end{equation*}

More simply, we seek 
$$\text{argmin}_\theta \mathbb{E}_{b_i \in B} \text{log}P(c_i | b_i; \theta).$$
\vspace{-.8cm}

\begin{figure}[h]
\vspace{-0.2cm}
    \centering
\begin{adjustbox}{width=.48\textwidth}
\footnotesize
\begin{tabular}{c}
\toprule
  Example Found Fix \\
\midrule
\begin{minipage}[t]{0.45\textwidth}
\begin{minted}[escapeinside=||]{java}
private double normalizeTime(double time,
            ec.graph.GraphInitializer init) {
    if (((init.maxTime) - (init.minTime)) == 0.0)
        return 1.0;
    else
        return |\colorbox{red}{((init.maxTime) - time)}| 
                / ((init.maxTime) - init.minTime);
}
\end{minted}
\end{minipage}\\
\midrule
\begin{minipage}[t]{0.45\textwidth}
\begin{minted}[escapeinside=||]{java}
private double normalizeTime(double time, 
            ec.graph.GraphInitializer init) {
    if (((init.maxTime) - (init.minTime)) == 0.0)
        return 1.0;
    else
        return |\colorbox{green}{(time - (init.minTime))}| 
                / ((init.maxTime) - init.minTime));
}
\end{minted}
\caption{A clever fix found by our model. It manipulates the return value so as to normalize \texttt{minTime} to 0 and \texttt{maxTime} to 1, whereas the developer originally had the boundary values reversed.}
\end{minipage}\\
\end{tabular}
\end{adjustbox}
\vspace{-0.6cm}
\label{fig:second_fix}
\end{figure}

\section{Data}
\par We evaluate on the Java dataset from \textit{Patches in the Wild} \cite{Tufano}. Tufano et al. mined all the public, non-fork repositories on Github up until 2016 for commits to Java code that include a commit message containing at least one of the following words: \textit{bug}, \textit{error}, \textit{issue} or \textit{fix}, \textit{patch}, \textit{correct}, which comprises about 5\% of all commits. From these commits they extracted all modified methods with at most one hundred tokens, and broke them into two datasets: a dataset of small methods containing less than 50 tokens and a complementary dataset of medium methods containing up to 100 tokens. 

\begin{figure*}[h]
\caption{Raw code, fully abstracted code, and code abstracted with idioms. We find that smaller models perform best on abstracted code, whereas higher-capacity models are better able to use the semantics associated to concrete code.}
\label{fig:task_example}
\centering
\includegraphics[scale=.3]{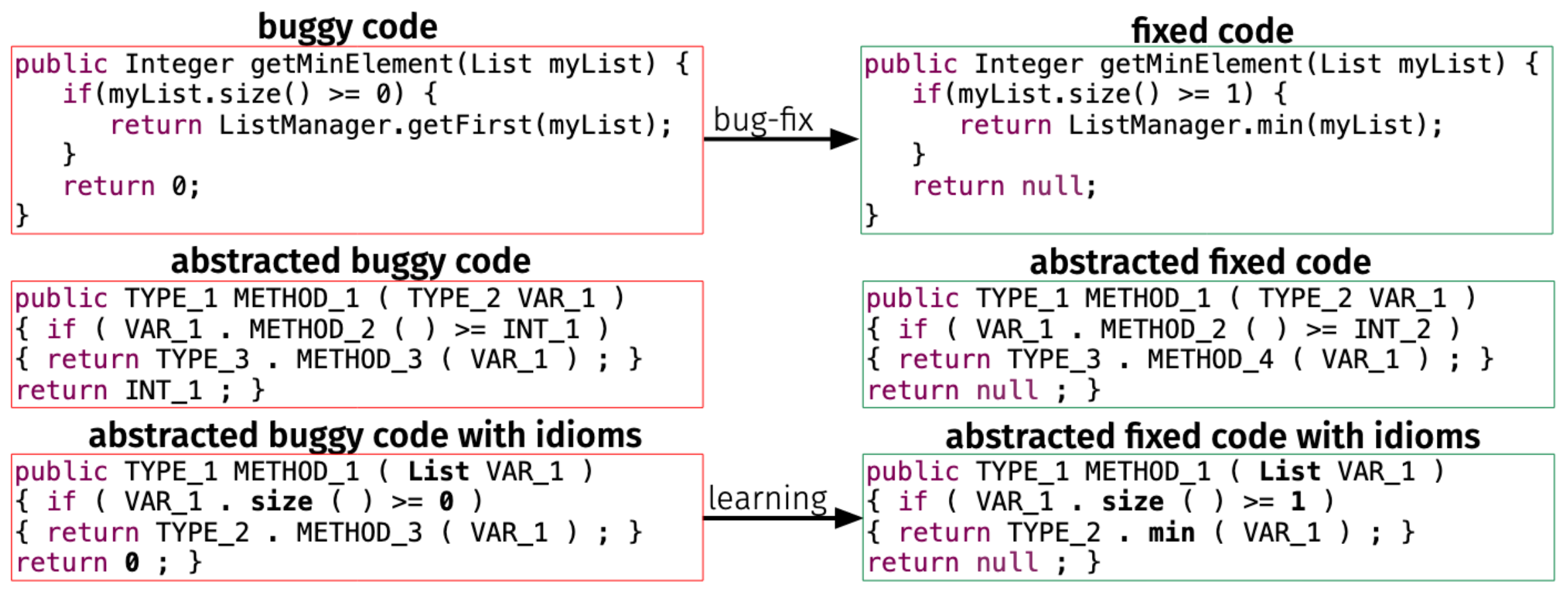}
\end{figure*}

To facilitate learning and to hasten convergence, Tufano et al. abstracted the code snippets while preserving necessary syntactical information. An example is given in figure \ref{fig:task_example}. To abstract the code snippets, they replaced all tokens outside a whitelist of approximately 500 commonly occurring tokens with abstractions of the form \texttt{METHOD}\textsubscript{i}, \texttt{VARIABLE}\textsubscript{i}, \texttt{STRING\_LIT}\textsubscript{i}, \texttt{NUM\_LIT}\textsubscript{i}, and \texttt{TYPE}\textsubscript{i}. Finally, they introduce \textit{idioms} -- frequently occurring programming language identifiers and literals, to which abstraction is not applied. Figure~\ref{fig:task_example} illustrates differences between these code representations.    
\par
As part of normalization the authors of the dataset also stripped comments and irrelevant whitespaces and indentation. In this paper, we present results both on the concrete code, and abstracted code with idioms. We obtain superior results on concrete code, especially after pretraining.

\section{Baseline}
Abstracting code removes the valuable semantics of variable and function names, and insisting on the verbatim fix is very strict, but it is nevertheless possible to obtain remarkably high accuracy under these conditions. Using a 10 million parameter bidirectional LSTM with copy mechanism and the default hyperparameters from Google's Seq2Seq Tufano et al. were able to get 9.2\% exact accuracy on small methods with less than 50 tokens, and 3.2\% accuracy on medium methods with between 50 and 100 tokens. Even more impressively, \textit{Copy That!} \cite{panthaplackel2020copy} got 17.7\% and 8.0\% accuracy respectively using their bi-GRU implementation with a novel, span-copying mechanism. 

We speculate that 17.7\% and 8.0\% are not far from the best practical top-1 accuracy on this task. ``Bug-fixing commits'' mined based on their commit message tend to be very diverse and noisy, in contrast to methods that are truly known to contain a bug, especially from a small class of synthetic bugs.

\section{Model}
Our \texttt{DeepDebug} model uses the standard transformer \cite{Vaswani} architecture. Copy mechanisms have been shown to give a dramatic boost to small LSTM models \cite{sequencer}, and to give a small boost to transformers for especially relevant tasks like summarization \cite{see2017get} and grammar correction \cite{Zhao}. We use copy-attention when training our 60 million parameter models from scratch and when replicating the state of the art. We do not use copy attention in our full-scale pretraining experiments, due primarily to technical limitations. 

We use a Byte Pair Encoding (BPE) vocabulary of roughly 50k tokens for concrete code and a vocabulary of all 433 unique tokens for abstracted code \cite{bpe}. In order to form the BPE vocabulary, we take the existing byte-level BPE tokenizer utilized by each of the GPT-2 \cite{gpt2}, BART, and RoBERTa, and append whitespace tokens in the order that they are learned by a tokenizer trained from scratch on raw code. Adding whitespace tokens effectively increases the context window length and processing speed by around 40\% during pretraining, although the improvement is smaller during finetuning as we restrict our focus to left-adjusted methods. Extending the pre-existing tokenizer allows us to reuse public checkpoints trained on English while efficiently adapting them to source code. 

\begin{figure*}[h!]
\caption{We experiment with adding syntax embeddings along with the standard positional and vocabulary embeddings during encoding. We find that this gives a small benefit when training from scratch, but no benefit when the syntax embeddings are added only during finetuning.
}
\centering
\includegraphics[width=0.7\textwidth]{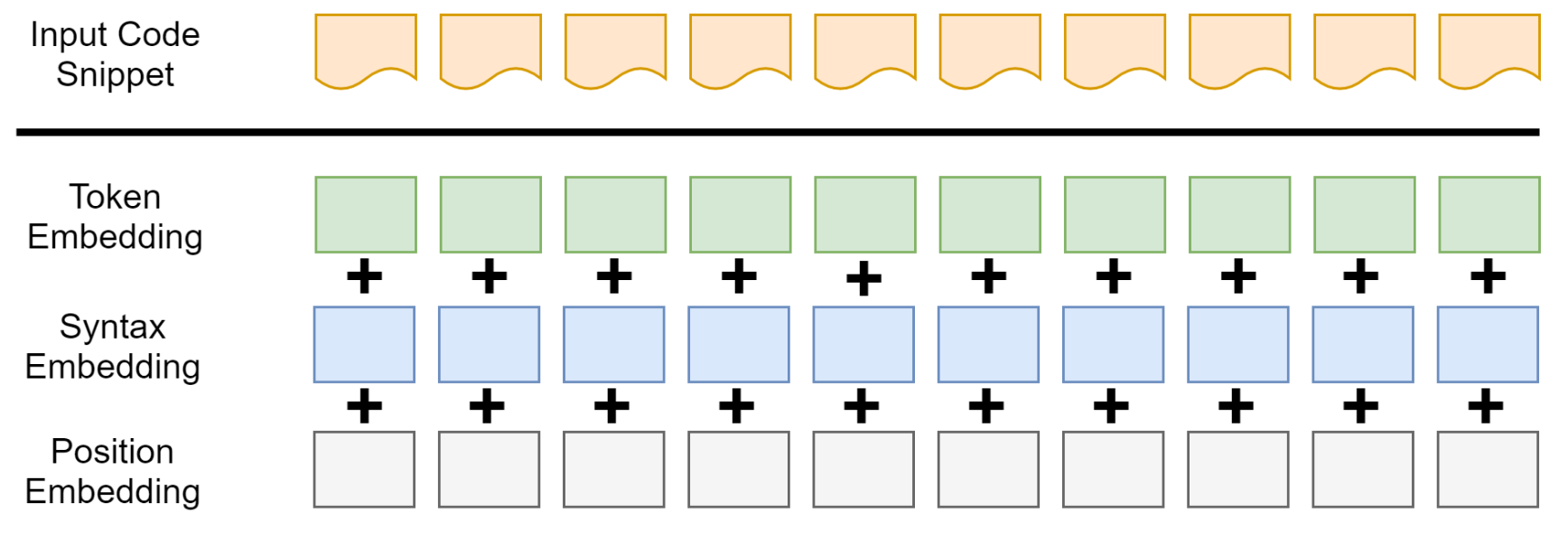}
\label{fig:syntax_embeddings}
\end{figure*}

We experiment with adding syntactic embeddings along with the standard vocabulary and positional embeddings, as demonstrated in figure \ref{fig:syntax_embeddings}. In this way we immediately distinguish whether BPE tokens like ``return'' or ``static'' are key words. In general we relieve the model of some of the difficulty of parsing the language's formal grammar, which is disproportionately difficult for transformers \cite{hahn2020theoretical}. We can also think of this tactic as subsuming both the approach of using raw code and the approach of using abstracted code. 

\section{Pretraining}
We use the same span-masking objective used by T5 \cite{T5}. That is, we randomly select spans of three consecutive tokens and replace each span with a single mask token. The goal is to reproduce the masked tokens in order, reusing the mask token in the target by putting it before every span of three tokens. We take advantage of the fact that code is lower-entropy than English. For instance, GPT-C attains perplexities in the range of 1.3 -- 2 for several programming languages, whereas a comparably sized GPT-2 model attains a perplexity of 26 on Wikitext-3 \cite{gpt-c,gpt2}. Thus we mask a larger fraction of 30\% of all tokens. Masking a larger fraction also has the added advantage of increasing the effective context length during pretraining, by about 50\% in our case.

\begin{figure}
    \centering
\begin{adjustbox}{width=.48\textwidth}
\footnotesize
\begin{tabular}{c}
\toprule
  Span-masking Input and Output \\
\midrule
\begin{minipage}[t]{0.45\textwidth}
\begin{minted}[escapeinside=||]{java}
private double normalizeTime(<mask>.GraphInitializer init) {
    if (((init.maxTime) - <mask>0.0)
        return 1.0;
    else
        <mask> (init.minTime)) / 
                ((init.maxTime) - (init.minTime));
    }
\end{minted}
\end{minipage}\\
\midrule
\begin{minipage}[t]{0.45\textwidth}
\begin{minted}[escapeinside=||]{java}
<mask>double time, ec.graph <mask> (init.minTime)) == 
<mask> return (time -
\end{minted}
\caption{Example of span-masking used when pretraining our sequence-to-sequence model. We reuse a single mask token and cover exactly three tokens with each mask.}
\end{minipage}\\
\end{tabular}
\end{adjustbox}
\label{fig:pub_priv_swap}
\end{figure}

We pretrain on all open-source Java repositories hosted on GitHub with at least 10 stars that have been updated in the last five years, 67,000 repositories in total. We deduplicate files based on file hash. Finally, we filter out files based on heuristics like if they contain data or have been auto-generated, and clean files by removing licenses and non-ASCII characters. This final stage of filtering and cleaning removes 10\% of the remaining tokens. After taking these steps, we are left with 67k repositories, 8 million files, 1.3 billion lines of code, 13 billion tokens, or 54GB.

We pretrain for ten epochs, which takes one week on a DGX-2 box, which has sixteen 32GB V100 GPUs. We estimate this is approximately 4\% of the resources used to pretrain the public English BART checkpoint released by Facebook, which was pretrained for forty epochs on 160GB using a more compute-intensive pretraining objective.


\section{Experiments}

We first examine training on concrete code as compared to abstracted code (with the 433 most common tokens added as idioms) from scratch using a 60 million parameter transformer with a copy mechanism as implemented in the OpenNMT framework \cite{OpenNMT}.

We then experiment with several pretraining strategies implemented in the Fairseq framework \cite{fairseq}. Specifically, we consider three following experiments: i) pretrain on Java only, ii) finetune directly from the strong baseline BART which was pretrained on English; and iii) domain-adaptive pretraining on Java with a warmstart from BART. We hypothesized that warmstarting from English would help, since learning how to program is much easier if one already knows English and has an intuitive understanding of concepts like dictionaries, graphs, or lists, and can read out these roles from self-documenting method and variable names.

\begin{figure*}[h]
\caption{We experiment with adding an auxiliary token-type classification loss for multitask training. In this condition, the loss is a weighted sum of the standard next-token prediction loss and the auxiliary loss of predicting whether each subword token is a part of a \texttt{TYPE}, \texttt{METHOD}, \texttt{VARIABLE}, \texttt{STRING\_LIT}, \texttt{NUM\_LIT}, or other token. 
}
\centering
\includegraphics[width=0.7\textwidth]{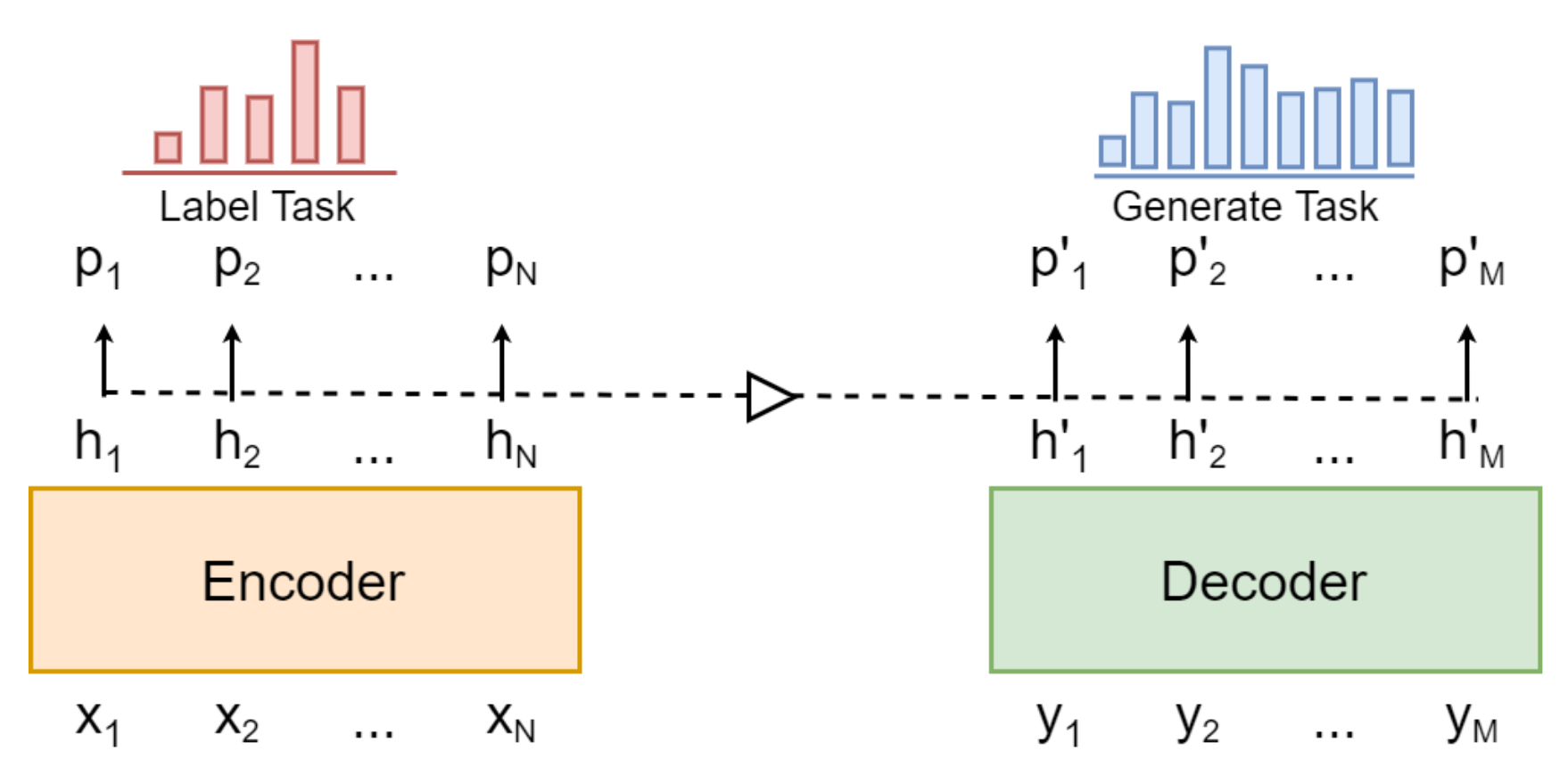}
\label{fig:token_labeling}
\end{figure*}

We next investigate the effect of adding an auxiliary token-type labeling loss on top of the standard next-token prediction loss, as demonstrated in Figure \ref{fig:token_labeling}. We use the same token types used during abstraction i.e. \texttt{METHOD}, \texttt{VARIABLE}, \texttt{STRING\_LIT}, and \texttt{NUM\_LIT}. Syntax highlighting is a ubiquitous developer tool, and especially helpful during the early stages of learning a language before the developer has internalized its formal grammar.

Our final experiment more closely investigates syntax highlighting. During encoding, we add a token-type embedding to the standard positional and vocabulary embeddings, as demonstrated in figure \ref{fig:syntax_embeddings}. For both of these token-type experiments we find a small improvement when training from scratch, but no improvement when adding them during finetuning only.
We also partially replicate and extend the work from \textit{Copy That!}, by evaluating on small methods on both concrete and fully abstracted code without idioms. We use the code for the specialized span-copying GRU model generously shared by the author.

\section{Results}
\label{sec:results}
We found that many of the fixes fell into a few uninspiring patterns, especially given the limited context of the methods: Simple deletions, as with classic models like GenProg; swapping ``protected'', ``private'', and ``public''; and inserting ``native''.

Indeed, roughly two-thirds of all the fixes we find on first attempt are simply deletions. For our strongest models, 7\% of the non-deletion fixes are some kind of protected/private/public swapping or inserting the word ``native'', while the fraction is as high as 50\% for the weakest models we consider. In presenting our results, we thus also give the number of fixes that were simply deletions and the complementary number that are non-deletion fixes.



\begin{table*}[h]
\centering
\caption{Transformer Results for Concrete vs. Abstract code. Contrary to the results for smaller GRU-based models, we see that training on concrete code helps.}
\resizebox{0.8\linewidth}{!}{
\begin{tabular}{lcccr}
\toprule
& Non-deletion Fixes & All Fixes & \multicolumn{1}{c}{All Methods} & \multicolumn{1}{c}{Deletion Fixes} \\ \midrule
\begin{tabular}[c]{@{}l@{}}
\bf{abstract with idioms medium},\\ from scratch, \\ 60M parameters   
\end{tabular} & 11 (0.2\%)              & 178 (2.7\%)       & \multicolumn{1}{c}{6545}        & \multicolumn{1}{c}{166}            \\ \hline
\begin{tabular}[c]{@{}l@{}}\bf{concrete medium},\\ from scratch,\\ 60M parameters\end{tabular} & \bf{20 (0.3\%)}               & \bf{244 (3.7\%)}       & \multicolumn{1}{c}{6545}        & \multicolumn{1}{c}{224}            \\ \hline
                                                              
\begin{tabular}[c]{@{}l@{}}
\bf{abstract with idioms small},\\ from scratch, \\ 60M parameters  \end{tabular}                                                       & 91 (1.6\%)              & 648 (11.1\%)       & \multicolumn{1}{c}{5835}        & \multicolumn{1}{c}{557}            \\ \hline
\begin{tabular}[c]{@{}l@{}}\bf{concrete small},\\ from scratch,\\ 60M parameters\end{tabular} & \bf{349 (6.0\%)}               & \bf{854 (14.6\%)}       & \multicolumn{1}{c}{5835}        & \multicolumn{1}{c}{505}            \\ \hline
\end{tabular}
}
\end{table*}

\begin{table*}[h]
\centering
\caption{GRU Span-copying Results for Small Java Methods, our replication of the previous state of the art \cite{panthaplackel2020copy}. Our pretrained model finds a slightly higher total number of fixes, and 75\% more non-deletion fixes.}\label{table:gru}
\resizebox{0.8\linewidth}{!}{
\begin{tabular}{lcccr}
\hline
                                                                        & Non-deletion Fixes & All Fixes & \multicolumn{1}{c}{All Methods} & \multicolumn{1}{c}{Deletion Fixes} \\ \hline
\bf{fully abstract} small,\\ from scratch,\\250K parameters                                                               & \bf{207 (3.5\%)}               & \bf{991 (17.0\%)}       & \multicolumn{1}{c}{5835}        & \multicolumn{1}{c}{784}            \\ \hline
\begin{tabular}[c]{@{}l@{}}\bf{concrete} small,\\ from scratch,\\250K parameters       \end{tabular} & 109 (1.5\%)               & 537 (9.2\%)       & \multicolumn{1}{c}{5835}        & \multicolumn{1}{c}{428}            \\ \hline
\end{tabular}
}
\end{table*}

\begin{table*}[h]
\centering
\caption{GRU Classical Copy Mechanism Results for Small Java Methods, model also provided by \cite{panthaplackel2020copy}. Unlike the transformers, the smaller, GRU-based models did not benefit from training on concrete code; on abstracted code they achieved both higher top-1 performance and a better ratio of non-deletion to deletion fixes.}
\resizebox{0.8\linewidth}{!}{
\begin{tabular}{lcccr}
\hline
                                                                        & Non-deletion Fixes & All Fixes & \multicolumn{1}{c}{All Methods} & \multicolumn{1}{c}{Deletion Fixes} \\ \hline
\bf{fully abstract} small,\\ from scratch,\\250K parameters                                                               & \bf{125 (2.1\%)}               & \bf{672 (11.5\%)}       & \multicolumn{1}{c}{5835}        & \multicolumn{1}{c}{547}            \\ \hline
\begin{tabular}[c]{@{}l@{}}\bf{concrete} small,\\ from scratch, \\250K parameters       \end{tabular} & 50 (0.9\%)              & 397 (6.8\%)       & \multicolumn{1}{c}{5835}        & \multicolumn{1}{c}{347}            \\ \hline
\end{tabular}
}
\end{table*}

\begin{figure}[h]
\vspace{-0.2cm}
    \centering
\begin{adjustbox}{width=.48\textwidth}
\footnotesize
\begin{tabular}{c}
\toprule
  Example Found Fix \\
\midrule
\begin{minipage}[t]{0.45\textwidth}
\begin{minted}[escapeinside=||]{java}
public void backward() {
    switch (heading) {
        case "N" :
            |\colorbox{red}{(x)++}|;
        default :
    }
}
\end{minted}
\end{minipage}\\
\midrule
\begin{minipage}[t]{0.45\textwidth}
\begin{minted}[escapeinside=||]{java}
public void backward() {
    switch (heading) {
        case "N" :
            |\colorbox{green}{(y)--}|;
            break;
        default :
            break;
    }
}
\end{minted}
\caption{Our model infers from the function name ``backward'' that the function should be decrementing rather than incrementing.}
\end{minipage}\\
\end{tabular}
\end{adjustbox}
\vspace{-1cm}
\label{fig:third_fix}
\end{figure}

When considering fixes that our model did not find, the overriding theme is that the task is impossible without more context. In theory, bug-fixing should be nearly deterministic; code can be edited in nearly arbitrary ways while implementing or optimizing additional functionality, but a bug is a glaring error that needs to be addressed. 

In practice, the correct behavior is often underspecified, especially since our approach excludes useful information such as test cases, example usages, and the class constructor. For example, consider the function pop() from figure ~\ref{fig:pop}, which our model did not fix on its first attempt. Off-by-one errors are common. However, given only the buggy method, it is ambiguous whether the head index should be bounded below by zero or negative one.

\begin{figure}[h]
\vspace{-0.2cm}
    \centering
\begin{adjustbox}{width=.48\textwidth}
\footnotesize
\begin{tabular}{c}
\toprule
  Example Fix not Found on First Attempt \\
\midrule
\begin{minipage}[t]{0.45\textwidth}
\begin{minted}[escapeinside=||]{java}
public G pop() {
    G output = storage[i_head];
    if ((i_head) |\colorbox{red}{>}| 0) {
        i_head = (i_head) - 1;
    }else {
    }
    return output
}
\end{minted}
\end{minipage}\\
\midrule
\begin{minipage}[t]{0.45\textwidth}
\begin{minted}[escapeinside=||]{java}
public G pop() {
    G output = storage[i_head];
    if ((i_head) |\colorbox{green}{>=}| 0) {
        i_head = (i_head) - 1;
    }else {
    }
    return output
}
\end{minted}
\caption{It is ambiguous whether this off-by-one error is a bug without seeing the definitions of storage or i\_head.}\label{fig:pop}
\end{minipage}\\
\end{tabular}
\end{adjustbox}
\vspace{-1cm}
\end{figure}

\section{Analysis}

In this section, we draw several conclusions from the results of the experiments presented in section~\ref{sec:results}.

First of all, we see that producing fixes for medium-sized methods is harder than for short methods. This is unsurprising since longer methods are more complicated and can be edited in more ways.

Our next observation is that pretraining helps, especially for medium methods. Indeed, the number of medium fixes found increases by a factor of three when adding pretraining.



We see that the smaller GRU models and the transformers trained on abstract code with idioms are especially destructive, producing seven times as many deletions as constructive fixes. The smaller GRU models benefit from abstracting code, although the larger transformers do not.

Pretraining on English boosts performance of program repair, presumably due to the semantics inherent in variable and function names. A function name like \texttt{getMinElement} is intuitive even for a non-programmer. Indeed, the public BART checkpoint, which was pretrained only on English, outperforms our model pretrained only on Java, although our Java-only model was trained for only 4\% as long as the English-only checkpoint. Our best results are obtained by taking the powerful English model and further pretraining it on Java.

We observe that adding an auxiliary syntax-labeling task helps somewhat when training from scratch, but makes no difference when we finetune the pretrained model. We hypothesize this is because the pretrained model already has a strong understanding of the language's syntax, and ample syntactical feedback is already provided by the standard next-token prediction task.

We see underwhelming returns in scaling from 60 million to 400 million parameters when training from scratch, especially in light of the dramatic and predictable gains from scaling up transformers such as T5 or GPT \cite{T5, LM_scaling}. We hypothesize that this is because of the small scale of training data. Each of our training sets contains only around 50k examples comprising mere tens of megabytes, whereas BART-large has 400 million parameters and is nearly a full gigabyte when loaded in FP16.

Finally, similarly to adding the syntax-labeling task, we see a small benefit from adding syntax highlighting in the form of adding token-type embeddings to the vocabulary embeddings during encoding, but no benefit from adding those embeddings during finetuning. It makes sense that there should be some small benefit from syntax highlighting. We hypothesize that the lack of benefit from adding highlighting during finetuning is due to the pretrained model already having developed a way to represent different token types, which clashes with the added embeddings.

\begin{table*}[p]
\centering
\caption{Transformer results for different pretraining workflows on medium methods. The best result is obtained by first pretraining on English and then on Java. Note that English pretraining used twenty-five times as much compute as Java pretraining.}
\resizebox{1.0\linewidth}{!}{
\begin{tabular}{lcccr}
\toprule
& Non-deletion Fixes & All Fixes & \multicolumn{1}{c}{All Methods} & \multicolumn{1}{c}{Deletion Fixes} \\ \midrule

\begin{tabular}[c]{@{}l@{}}concrete medium,\\ \bf{from scratch}, \\ 400M parameters\end{tabular} & 72 (1.1\%)              & 233 (3.6\%)       & \multicolumn{1}{c}{6545}        & \multicolumn{1}{c}{161}            \\ \hline
\begin{tabular}[c]{@{}l@{}}concrete medium,\\ \bf{Java pretrain only}, \\400M parameters\end{tabular} & 98 (1.5\%)               & 413 (6.3\%)       & \multicolumn{1}{c}{6545}        & \multicolumn{1}{c}{315}            \\ \hline

\begin{tabular}[c]{@{}l@{}}concrete medium,\\ \bf{English pretrain only}, \\400M parameters\end{tabular} & 143 (2.2\%)               & 439 (6.7\%)       & \multicolumn{1}{c}{6545}        & \multicolumn{1}{c}{296}            \\ \hline

\begin{tabular}[c]{@{}l@{}}concrete medium,\\ \bf{English plus Java}, \\ 400M parameters\end{tabular} & \bf{259 (4.0\%)}               & \bf{749 (11.4\%)}       & \multicolumn{1}{c}{6545}        & \multicolumn{1}{c}{490}            \\ \hline

\end{tabular}
}
\end{table*}

\begin{table*}[p]
\centering
\caption{Transformer results for different pretraining workflows on small methods. The English BART baseline is very strong, although the best performance on the standard top-1 accuracy metric is still obtained by further pretraining on Java.}
\resizebox{1.0\linewidth}{!}{
\begin{tabular}{lcccr}
\toprule
& Non-deletion Fixes & All Fixes & \multicolumn{1}{c}{All Methods} & \multicolumn{1}{c}{Deletion Fixes} \\ \midrule

\begin{tabular}[c]{@{}l@{}}concrete small,\\ \bf{from scratch},\\ 400M parameters\end{tabular}     & 328 (5.6\%)             & 816 (13.9\%)       & \multicolumn{1}{c}{5835}        & \multicolumn{1}{c}{488}            \\ \hline
\begin{tabular}[c]{@{}l@{}}concrete small,\\ \bf{Java pretrain only},\\ 400M parameters\end{tabular}     & 330 (5.7\%)              & 979 (16.8\%)       & \multicolumn{1}{c}{5835}        & \multicolumn{1}{c}{649}            \\ \hline
\begin{tabular}[c]{@{}l@{}}concrete small,\\ \bf{English pretrain only},\\ 400M parameters\end{tabular}     & \bf{385 (6.6\%)}              & 976 (16.7\%)       & \multicolumn{1}{c}{5835}        & \multicolumn{1}{c}{591}            \\ \hline
\begin{tabular}[c]{@{}l@{}}concrete small,\\ \bf{English plus Java},\\ 400M parameters\end{tabular}     & 361 (6.2\%)              & \bf{1090 (18.7\%)}       & \multicolumn{1}{c}{5835}        & \multicolumn{1}{c}{729}            \\ \hline
\end{tabular}
}
\end{table*}

\begin{table*}[p]
\centering
\caption{Effects of adding a syntax-type labeling task and of adding syntax-type embeddings during encoding. There is a small improvement for multitask training from scratch, but no improvement for adding the additional task during finetuning, presumably because the labeling task is too easy for the pretrained model. There is similarly a small benefit from adding syntax embeddings when training from scratch, perhaps the lack of benefit during finetuning is because the added embeddings clash with the pretrained model's representation of different token types.}
\resizebox{1.0\linewidth}{!}{
\begin{tabular}{lcccr}
\hline
                                                                        & Non-deletion Fixes & All Fixes & \multicolumn{1}{c}{All Methods} & \multicolumn{1}{c}{Deletion Fixes} \\ \hline
\begin{tabular}[c]{@{}l@{}}concrete small,\\ \bf{from scratch},\\ 400M parameters\end{tabular} & 328 (5.6\%)              & 816 (13.9\%)       & \multicolumn{1}{c}{5835}        & \multicolumn{1}{c}{488}            \\ \hline
\begin{tabular}[c]{@{}l@{}}concrete small,\\ \bf{from scratch},\\ 400M parameters, \\ \bf{syntax token embedding} \end{tabular} & \bf{343 (5.9\%)}           & 847 (14.4\%)       & \multicolumn{1}{c}{5835}        & \multicolumn{1}{c}{504}            \\ \hline
\begin{tabular}[c]{@{}l@{}}
concrete small,\\ \bf{from scratch}, \\ 400M parameters, \\ \bf{syntax token prediction}
\end{tabular}                                                       & 340 (5.8\%)               & \bf{862 (14.7\%)}       & \multicolumn{1}{c}{5835}        & \multicolumn{1}{c}{522}            \\ \hline
\begin{tabular}[c]{@{}l@{}}
concrete small,\\ \bf{English plus Java}, \\ 400M parameters,
\end{tabular}                                                       & \bf{361 (6.2\%)}               & 1090 (18.7\%)       & \multicolumn{1}{c}{5835}        & \multicolumn{1}{c}{729}            \\ \hline
\begin{tabular}[c]{@{}l@{}}
concrete small,\\ \bf{English plus Java}, \\ 400M parameters, \\ \bf{syntax token embedding}
\end{tabular}                                                       & 355 (6.1\%)               & 1082 (18.5\%)       & \multicolumn{1}{c}{5835}        & \multicolumn{1}{c}{727}            \\ \hline
\begin{tabular}[c]{@{}l@{}}
concrete small,\\ \bf{English plus Java}, \\ 400M parameters, \\ \bf{syntax token prediction}
\end{tabular}                                                       & 357 (6.1\%)              & \bf{1097 (18.8\%)}       & \multicolumn{1}{c}{5835}        & \multicolumn{1}{c}{740}            \\ \hline

\end{tabular}
}
\end{table*}

\begin{table*}[p]
\centering
\caption{Comparison bewteen prior SOTA results and our best model, which was warmstarted from BART and then further pretrained on Java. In table \ref{table:gru} we partially replicate Panthaplackel et al.'s small copy-span model and find that they only produce 57\% as many non-deletion fixes as our approach.} 
\resizebox{1.0\linewidth}{!}{
\begin{tabular}{lcccr}
\toprule
& Medium Fixes & Small Fixes & Overall & Parameters \\ \midrule

\begin{tabular}[c]{@{}l@{}}\bf{Tufano et al.}\\ \cite{Tufano} \end{tabular}     &  209 (3.2\%)             & 537 (9.2\%)    & 746 (6.0\%)   & \multicolumn{1}{c}{10M}                   \\ \hline

\begin{tabular}[c]{@{}l@{}}\bf{Panthaplackel et al.}\\ \cite{panthaplackel2020copy} \end{tabular}     & 524 (8.0\%)            & 1033 (17.7\%)    & 1557 (12.6\%)   & \multicolumn{1}{c}{0.25M}                   \\ \hline

\begin{tabular}[c]{@{}l@{}}\bf{CodeBERT}\\ \cite{codebert} \end{tabular}     & 340 (5.2\%)             & 957 (16.4\%)    & 1297 (10.5\%)   & \multicolumn{1}{c}{180M}                   \\ \hline


\begin{tabular}[c]{@{}l@{}}\bf{BART}\\ \cite{BART} \end{tabular}     & 439 (6.7\%)             & 976 (16.7\%)    & 1415 (11.4\%)   & \multicolumn{1}{c}{400M}                   \\ \hline

\begin{tabular}[c]{@{}l@{}}\texttt{DeepDebug}\\ (Ours) \end{tabular}     & \bf{749 (11.4\%)}             & \bf{1090 (18.7\%)}    & \bf{1839 (14.9\%)}   & \multicolumn{1}{c}{400M}                   \\ \hline
\end{tabular}
}
\end{table*}

\section{Conclusions and Future Work}
We introduced \texttt{DeepDebug}, a data-driven program repair approach which learns to fix bugs in real-world code using pretrained transformers. We achieve new state of the art by following a multi-stage pretraining pipeline, in which we warmstart from a public English checkpoint, further pretrain on code, and finally finetune on the target task. We show that these large transformers are able to take advantage of meaningful function and variable names, unlike smaller GRU models which perform better on abstracted code. We also investigate the extent to which our fixes are constructive, rather than mere deletions, and find that we create 75\% more non-deletion fixes than the previous state of the art.

In future work we are interested in expanding to multiple languages and considering iterative fixes.





\bibliographystyle{acl_natbib}
\bibliography{bugpatching}

\end{document}